\documentclass[11pt]{article}

\usepackage{acl}

\usepackage{times}
\usepackage{latexsym}

\usepackage[T1]{fontenc}

\usepackage[utf8]{inputenc}

\usepackage{microtype}

\usepackage{inconsolata}

\usepackage{graphicx}
\usepackage{amsmath}
\usepackage{amsfonts}
\usepackage{multirow}
\usepackage{booktabs}

\title{Supplement Generation Training for Enhancing Agentic Task Performance}

\author{
  Young-Min Cho$^\dag$\thanks{~~Work done during internship at Amazon.}\quad Daniele Bonadiman$^\ddag$ \quad Divya Bhargavi$^\ddag$ \\[0.3em]
  {\bf Tamer Alkhouli$^\ddag$ \quad Salvatore Romeo$^\ddag$ \quad Dongwei Jiang$^\ddag$ \quad Khushbu Pahwa$^\ddag$} \\[0.3em] { \bf  Yubin Ge$^\ddag$ \quad Etsuko Ishii$^\ddag$ \quad Monica Sunkara\footnotemark[4] \quad Yi Zhang$^\ddag$} \\[0.8em]
  $^\dag$University of Pennsylvania, $^\ddag$ AWS Agentic AI Labs \\[0.3em]
  {\texttt{jch0@seas.upenn.edu, \{dbonadim, yizhngn\}@amazon.com}}
}

\begin{document}
\maketitle
\footnotetext[4]{~~Work done while at Amazon.}

\begin{abstract}
Training large foundation models for agentic tasks is increasingly impractical due to the high computational costs, long iteration cycles, and rapid obsolescence as new models are continuously released. Instead of post-training massive models for every new task or domain, we propose Supplement Generation Training (SGT), a more efficient and sustainable strategy. SGT trains a smaller LLM to generate useful supplemental text that, when appended to the original input, helps the larger LLM solve the task more effectively. These lightweight models can dynamically adapt supplements to task requirements, improving performance without modifying the underlying large models. This approach decouples task-specific optimization from large foundation models and enables more flexible, cost-effective deployment of LLM-powered agents in real-world applications.
\end{abstract}

\section{Introduction}
\begin{figure}[t]
    \centering
    \includegraphics[width=\linewidth]{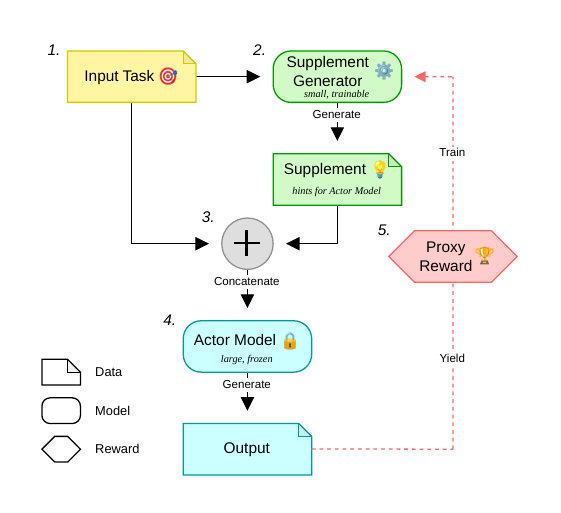}
    \caption{Overview of Supplement Generation Training. Instead of giving the input task directly to the actor model, it is first routed through the supplement generator, which produces supplemental information to aid in task completion. This supplement is concatenated with the original input and passed to the actor model, which solves the task with the additional guidance. The quality of the actor's solution then serves as a proxy reward for training the supplement generator, indirectly signaling the effectiveness of the generated supplement.}
    \label{fig:spirit}
\end{figure}

The most capable language models today are increasingly deployed as closed-source, API-only systems, where gradient access is unavailable. Even when fine-tuning remains an option, the computational overhead is significant, and newly released foundation models routinely render task-specific fine-tunes obsolete before they can be fully amortized. For these reasons, improving model behavior without modifying weights is often the only viable path, and optimization pressure naturally shifts from the model to the input \citep{cheng2024bpo, zhou2022ape}.

Prompt optimization is the natural response to this constraint, but most current methods primarily select from or rearrange existing artifacts rather than generating new, input-specific content. Global methods such as DSPy and TextGrad optimize instruction templates or textual variables across an entire dataset \citep{khattab2023dspy, yuksekgonul2025textgrad}. Recent query-dependent approaches such as Local Prompt Optimization and Prompt-OIRL tailor prompts to individual inputs \citep{jain2025local, sun2023query}, yet their adaptation largely operates within a fixed pool of templates. Whether the search is global or per-instance, these methods generally do not learn to synthesize new reasoning structures that may be necessary to solve the task.

The dynamic of training and prompt optimization are similar to the relationship between an executive and their assistant. The executive is highly capable but expensive and fully occupied; they cannot be retrained for every new task. The assistant's role is not merely to relay instructions verbatim but to prepare the right context, provide relevant background, and frame each problem so that the executive can apply their expertise most effectively. A good assistant learns, over time, which framings lead to better decisions, what supplementary information proves useful, and how to anticipate the executive's reasoning process. In much the same way, we argue that prompt optimization should not be limited to selecting or rearranging existing templates but should learn to synthesize new supporting content tailored to each input, preparing the context a frozen model needs to perform at its best.

Inspired by this relationship, we propose \textbf{Supplement Generation Training (SGT)}. Trained with proxy rewards from the task outcomes, SGT teaches a small, open-source model to generate short, instance-specific textual supplements that steer a frozen actor at inference time. Given a query, the \textit{supplement generator} produces an instance-specific supplement such as a summary, a set of likely mistakes, or a reasoning scaffold, which is appended to the input before the actor is called. With the additional context provided by the supplement generator, the \textit{actor model} can solve the task more effectively, without any modification to its weights.

SGT offers several compelling advantages. First, it is highly cost-efficient: rather than relying on hundred-billion-parameter models, SGT primarily trains a 1.7B model that is lightweight and fast to train, dramatically reducing computational overhead. Second, SGT dynamically selects the supplement type based on the specific task, input instance, and target actor model, enabling targeted assistance rather than one-size-fits-all augmentation. Third, SGT is broadly generalizable, demonstrating consistent improvements across a wide range of benchmarks, actor LLMs, and supplement generator LLMs.

In our experiments, we show that SGT outperforms baseline models without supplements by 21\%, showing consistent improvements across a wide range of benchmarks and LLMs from different model families. Through ablation studies, we further demonstrate the critical role of iterative preference optimization in gradually achieving these gains. In the qualitative analysis, we further examine the distribution of supplement types across different training stages and reveal a search-and-focus strategy emerging throughout the supplement sampling process.

\noindent
Our work makes the following key contributions:
\begin{itemize}
    \item \textbf{Supplement Generation Training (SGT).} We introduce a new framework that trains a small \emph{supplement generator} model to produce per-input textual supplements that enhance the performance of large LLMs without modifying their parameters.
    
    \item \textbf{Efficient training pipeline with SFT+DPO.} We develop a scalable training procedure combining warm-start SFT and iterative DPO, leveraging outcome-based rewards from a black-box actor model.
    
    \item \textbf{Comprehensive empirical validation.} We evaluate SGT across 5 benchmarks, achieving 21\% improvement over baselines. Further analysis reveals strong generalization across model families and tasks.
\end{itemize}

\begin{figure*}[t]
    \centering
    \includegraphics[width=\textwidth]{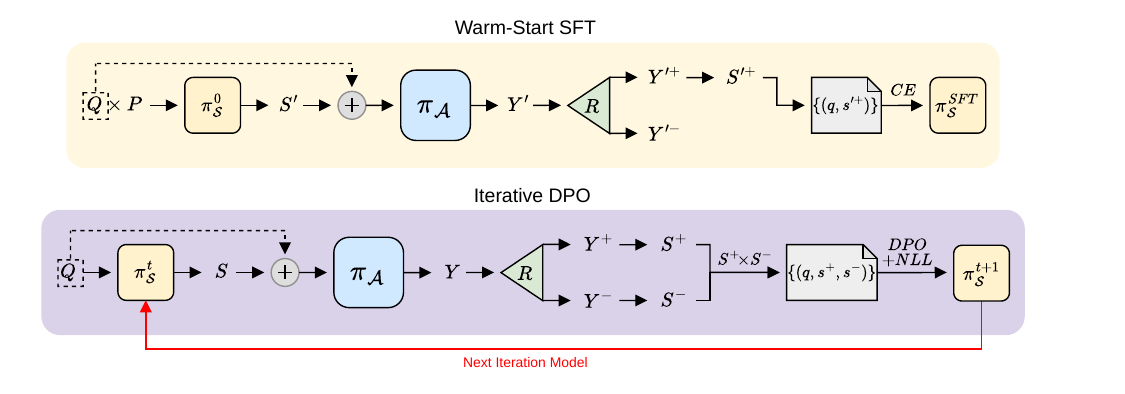}
    \caption{Illustration of our proposed training pipeline: Supplement Generation Training. To teach model how to generate, how to select, and how to improve supplements, we structure the pipeline into a Warm-Start SFT phase followed by an iterative DPO phase.}
    \label{fig:training}
\end{figure*}

\section{Related Works}
Automatic prompt optimization (APO) aims at replacing manual prompt tuning with algorithms that search instructions or exemplars. Global/template methods include LLM-driven instruction induction and reinforcement learning over discrete prompt tokens \citep{zhou2022ape,deng2022rlprompt}. Black-Box Prompt Optimization (BPO) adapts this to API-only models by iteratively editing prompts from task-level or preference feedback \citep{cheng2024bpo}. Other effective global/template methods that emerged in recent years utilize textual reflection-based feedback to optimize the prompts \citep{yuksekgonul2025textgrad,agrawal2025gepa}. DSPy introduces a programming abstraction for LLM pipelines, compiling declarative modules into optimized prompts by automatically searching over instruction and demonstration configurations \citep{khattab2023dspy}. TextGrad treats LLM-generated feedback as textual gradients and applies gradient-descent-like updates to iteratively refine prompt variables \citep{yuksekgonul2025textgrad}. While both methods effectively optimize global prompt templates, they do not generate new, instance-conditioned content tailored to individual inputs. Beyond global template search, recent work formalizes \emph{local} and \emph{query-dependent} automatic prompt optimization, where prompts are tailored per input rather than fixed across a dataset. Local Prompt Optimization (LPO) constrains editing to salient ``optimization tokens,'' reporting faster convergence and stronger gains than global methods on reasoning benchmarks \citep{jain2025local}. In parallel, Prompt-OIRL frames \emph{query-dependent} prompt selection via an offline reward model learned from historical prompt--outcome logs, enabling per-input prompt choice without calling the target LLM at inference time \citep{sun2023query}. These methods optimize or select the \emph{prompt} itself, often by editing an instruction template, scoring candidate prompts with an offline evaluator, or tuning prompts across test samples, whereas our approach keeps any base template unchanged and appends a short, generated, instance-conditioned supplement (e.g., reasoning seed, summary, rephrase, contrastive cue) to the input at inference time. In addition, query-dependent APO frequently relies on offline reward models to score/edit prompts \citep{sun2023query,kong2024qpo}, while our learning signal is the task-outcome as a proxy reward signal returned by a frozen, API-accessed actor.

Several lines of work deploy a \emph{separate} model at inference time to improve a stronger model by altering its \emph{inputs}. \cite{liu2022generatedknowle} shows that LLM-generated contextual knowledge can improve the performance of an actor model. Instance-level example selection learns a retriever that picks in-context demonstrations tailored to each query, yielding consistent gains across tasks and target LLMs \citep{rubin2021learning}. Complementarily, prompt compression models summarize or prune long contexts before the actor call, improving accuracy and latency in long-context settings while mitigating position effects \citep{jiang2023llmlingua,jiang2023longllmlingua}. 

Together these approaches instantiate the pattern we study: a small, dedicated model operates \emph{pre-inference} to construct a better input for a larger, frozen actor. Our work is similar in spirit but differs in a crucial way: in this work, we are not optimizing for a specific supplement type but rather we aim at identifying a training recipe for supplement generation models that adapt and evolve based on task specific signals.

\section{Methodology}
We consider a framework where a supplement generator LLM $\pi_\mathcal{S}$ is trained to produce a supplementary text $s$ given an input query $q\in Q$. This supplement is then combined with the query and provided to a frozen actor LLM $\pi_\mathcal{A}$, which generates the final output $y=\pi_\mathcal{A}(q,s)$. Crucially, $\pi_\mathcal{S}$ is not responsible for solving the task directly; instead, it learns to generate supplements that guide the frozen actor LLM $\pi_\mathcal{A}$ to produce improved outputs. The pipeline is illustrated in Figure \ref{fig:spirit}.

In this section, we first define the concept of a supplement, then describe the Proxy Reward Signal, and finally present our proposed training pipeline: Supplement Generation Training.

\subsection{Definition of Supplement}
\label{sec:supp}
A \textit{supplement} $s$ is an auxiliary piece of information generated to enhance or complete an input (e.g., a question, prompt, or task). It provides additional context, clarification, reasoning, or structure that helps the model or agent perform more effectively in subsequent reasoning or decision-making steps. While the type and format of a supplement can be diverse, in this work, we pre-define eight supplement types based on commonly used ones in previous literature:

 \textit{1. Answer}: Deliver a direct response to the problem, akin to a Multi-Agent Debate setting where others' answers are visible for comparison \cite{du2023improving}.
 
 \textit{2. Background}: Supply supplementary context or external knowledge related to the input \cite{liu2022generatedknowle}.
 
 \textit{3. Chain-of-Thought (CoT)}: Offer step-by-step reasoning or planning underlying the solution \cite{wei2022chainofthought}.
 
 \textit{4. Rephrase}: Reformulate the task to present alternative perspectives and reduce ambiguity \cite{chen-etal-2024-self-para}.
 
 \textit{5. Summary}: Present key information upfront to minimize cognitive effort \cite{liu2024lostinthemiddle,jiang2023longllmlingua}.
 
 \textit{6. Mistakes}: Warn about frequent or simple errors that could occur \cite{zhang2024context}.
 
 \textit{7. One-shot}: Generate a synthetic single example to guide task completion \cite{brown2020gpt3}.
 
 \textit{8. Pairs}: Provide contrasting correct and incorrect examples to highlight potential pitfalls \cite{gao2024contrastive}.

Note that the predefined list of supplement types is not exhaustive. Users may revise or expand the list based on their understanding of the task or their preferred supplement generation behavior. This set of types serves as the initial configuration that guides the LLM’s supplement generation process.

\subsection{Proxy Reward Signal}
\label{sec:PRS}
The quality of a supplement $s$ is inherently subjective and difficult to define. In this work, we leverage the reward signal obtained from the final output of the actor model $\pi_\mathcal{A}$ to indirectly reflect the quality of the paired supplement $s$.

Training begins with a dataset consisting of queries $Q=\{q_1, \ldots, q_n\}$ and their corresponding gold-standard responses $Y^*=\{y_1^*, \ldots, y_n^*\}$. Each query $q \in Q$ contains all necessary information to solve the task, such as the question, background, and instruction, while $Y^*$ is later used for reward evaluation. A reward model $R:(y, y^*) \rightarrow \mathbb{R}$ is defined to assess the quality of a generated output $y$. For simplicity, we constrain the rewards $r = R(y, y^*)$ to binary values, indicating whether the task was successfully solved.

For each query $q$, a supplement $s$ is associated with a unique output $y$ determined by $y = \pi_\mathcal{A}(q, s)$. Hence, we use the reward assigned to $y$ as a proxy indicator of the quality of $s$. Given $q$, $S$, and $y^*$, the reward model $R$ partitions the supplement set $S$ into a positive set $S^+ = \{\,s \in S \mid R(y, y^*) = 1,\, y = \pi_\mathcal{A}(q, s)\,\}$ and a negative set $S^- = \{\,s \in S \mid R(y, y^*) = 0,\, y = \pi_\mathcal{A}(q, s)\,\}$.

\subsection{Supplement Generation Training}
Building on the definitions above, we introduce our training framework, \textbf{Supplement Generation Training (SGT)}. The goal of this framework is to teach the model: 1) how to generate a supplement, 2) how to select the appropriate type of supplement, and 3) how to improve the quality of a supplement. To achieve these goals progressively, we structure the pipeline into a Warm-Start SFT phase followed by an iterative DPO phase.

\subsubsection{Warm-Start SFT}
The objective of the Warm-Start SFT phase is to train the model to produce supplements in the correct format while learning to select from the predefined eight supplement types. Since an untrained LLM naturally focuses on solving the input task rather than producing supplements, jumping straight into the DPO phase would be challenging, as the model’s initial behavior differs significantly from the desired one. The warm-start SFT stage bridges this gap, enabling the model to align with the target behavior and ensuring a smoother, more effective DPO training process.

To build the SFT training dataset, we begin by prompting the untrained model $\pi^0_\mathcal{S}$ with each of the eight predefined supplement types, along with an additional \textit{Free Style} type.\footnote{For details of the prompt and supplement format, see Appendix \ref{supp_types}.} For each input query $q$, this process yields nine generated supplements. We repeated (sampled) the process 5 times for all queries to form a prompt-guided supplement set $S'$. Each supplement $s \in S'$ is then used to produce an output $y = \pi_\mathcal{A}(q, s)$. As detailed in Section~\ref{sec:PRS}, we employ the Proxy Reward Signal to identify a positive subset of supplements, $S'^+$, that led $\pi_\mathcal{A}$ to successful outcomes. After stratified sampling on the supplement type, we then pair these successful supplements with their corresponding queries to form the dataset $\{(q, s'^+)\}$, which is used to perform SFT training of $\pi^0_\mathcal{S}$ using Cross-Entropy loss, resulting in the updated model $\pi^1_\mathcal{S}$. 

\subsubsection{Iterative DPO}
The objective of the iterative DPO phase is to identify the most effective supplement types, including previously unseen ones, and to enhance the overall quality of generated supplements. This motivates an iterative training paradigm in which, at each iteration $t$, the current model $\pi^t_\mathcal{S}$ generates new training data that is subsequently used to train the next model $\pi^{t+1}_\mathcal{S}$~\cite{pang2024iterative}, until reaching the maximum iteration $T$. 

This stage operates through a form of natural selection: supplements that lead the actor to successful outcomes are more frequently placed on the chosen side of DPO pairs, shifting the generator's output distribution toward more effective strategies. The updated generator then samples the data for the next iteration, allowing the pipeline to progressively refine both the selection of supplement types and the quality of supplements within them.

During the early iterations, the process focuses on expanding the search space by exploring diverse supplement types, whereas later iterations emphasize refinement and optimization within a narrower space. This \textit{search-and-focus} strategy is achieved by varying the sampling methods used to construct the supplement set $S$.

In the first iteration of DPO, the supplement set $S$ is constructed from three distinct sources for each given query:\\ 
\begin{enumerate}
    \item \textbf{Pre-Defined:} The eight pre-defined supplement types described in Section~\ref{sec:supp}.
    
    \item  \textbf{Out-of-Distribution (OOD):} The three most probable supplement types that do not belong to the pre-defined set.
    
    \item  \textbf{Concatenated (Concat):} Three pairs of successful supplement types.
\end{enumerate}

In this sampling strategy, generating specific supplement types can be enforced via an LLM output prefix, while the probability of a supplement type is estimated from the token probabilities corresponding to its indicator position within the pre-defined supplement format.\footnote{For details of the supplement format, see Appendix \ref{supp_types}.}. We sample 5 times to construct the final $S$ for the first iteration. In subsequent iterations, we simply sample 20 times from $\pi^t_\mathcal{S}$ without enforcing specific supplement types to construct $S$.

After constructing the supplement set $S$, it is divided into $S^+$ and $S^-$ using the Proxy Reward Signals described in Section~\ref{sec:PRS}. When forming preference pairs from these two subsets, two types of pairs can be distinguished:\\ 

    \textbf{Cross-Type Pairs}: Pairs composed of supplements belonging to different types. These pairs guide the model in selecting preferable supplement types. When the chosen type is \textit{OOD}, the model is encouraged to explore new supplement categories; when the chosen type is \textit{Concat}, it incentivizes the generation of multiple diverse supplements.
    
    \textbf{Within-Type Pairs}: Pairs composed of supplements of the same type. Since no type distinction exists, these pairs focus on teaching the model how to generate higher-quality supplements within a specific type.\\ 

To mitigate potential bias toward specific types of supplement pairs, we apply a cap of 20 pairs per pair type and perform stratified sampling based on the chosen supplement types within each category of pairs. 

Using the resulting sampled supplement pairs, we train the model $\pi^t_\mathcal{S}$ with a combination of DPO loss and Negative Log-Likelihood (NLL) loss, as the NLL term has been shown to play an essential role in DPO training~\cite{pang2024iterative}:
\[
\mathcal{L} = \mathcal{L}_{\text{DPO}} + \alpha \mathcal{L}_{\text{NLL}}
\]
The NLL loss is length-normalized, and we set $\alpha = 1$ following the configuration used in \citealp{pang2024iterative}.

\section{Experiments}

\subsection{Benchmarks}

We evaluate the effectiveness of the proposed SGT pipeline on five benchmarks covering \textit{structured output}, \textit{code generation}, \textit{multi-hop reasoning}, and \textit{frontier knowledge}. These benchmarks were selected to provide a representative and broadly applicable range of task types.\\

     \textbf{Spider} \cite{yu2018spider}: A complex text-to-SQL benchmark that evaluates a model’s ability to generate structured database queries from natural language questions across unseen schemas.
     
     \textbf{DS-1000} \cite{lai2023ds}: A code generation benchmark derived from real-world Jupyter notebooks, testing models on data science tasks involving libraries such as NumPy and pandas.
     
     \textbf{HotpotQA} \cite{yang2018hotpotqa}: A multi-hop question answering dataset requiring reasoning over multiple documents to synthesize supporting facts for complex questions. We provide only gold paragraphs to focus on evaluating multi-hop reasoning ability.
     
     \textbf{Humanity's Last Exam (HLE)} \cite{phan2025humanity}: A frontier knowledge benchmark composed of expert-authored academic questions across diverse disciplines to assess deep conceptual reasoning. We use the text-only subset in our experiments.
     
     \textbf{superGPQA} \cite{du2025supergpqa}: A large-scale graduate-level QA benchmark spanning hundreds of professional fields, designed to evaluate advanced domain-specific expertise and reasoning.\\

The creation of large, high-quality datasets remains difficult, primarily because of annotation inaccuracies and the considerable time and expense required for human annotation \cite{liu2024automatic}. To better reflect real-world conditions and highlight the efficiency of SGT in low-resource settings, we intentionally select smaller-scale datasets or down-sample from the original instances. A summary of benchmark statistics is provided in Table~\ref{tab:benchmark_sizes}.

\subsection{Baselines}
We compare the performance of our method with a wide range of baselines and ablations\footnote{Additional ablation studies are presented in Appendix \ref{app:dpo_iter}, \ref{app:model_size} and \ref{app:cross_dataset}.}:

\begin{table*}[t]
    \centering
    \resizebox{\textwidth}{!}{
        {\renewcommand{\arraystretch}{1.3}
            \begin{tabular}{lccccccccccc}
                \toprule
                \multirow{2}{*}{\textbf{Method}} & \multicolumn{5}{c}{\textbf{v3.5-sonnet-v2}} & \multicolumn{5}{c}{\textbf{gpt-oss-120b}} & \multirow{2}{*}{\textbf{Avg. Gain}} \\
                \cmidrule(lr){2-6} \cmidrule(lr){7-11}
                & \textbf{Spider} & \textbf{DS-1000} & \textbf{HotpotQA} & \textbf{HLE} & \textbf{superGPQA} & \textbf{Spider} & \textbf{DS-1000} & \textbf{HotpotQA} & \textbf{HLE} & \textbf{superGPQA} & \\
                \midrule
                $\emptyset\rightarrow\pi_\mathcal{A}$                    & 0.641 & 0.565 & 0.690 & 0.035 & 0.203 & 0.707 & 0.540 & 0.697 & 0.025 & 0.372 & -- \\
                $\emptyset\rightarrow\pi^{ITS}_\mathcal{A}$              & 0.647 & 0.580 & 0.647 & 0.019 & \textbf{0.295} & 0.704 & 0.550 & 0.663 & 0.037 & 0.385 & 5\% \\
                $\emptyset\rightarrow\pi^{solve}_\mathcal{S}$            & 0.495 & 0.200 & 0.573 & 0.046 & 0.092 & 0.495 & 0.200 & 0.573 & 0.046 & 0.092 & -23\% \\
                $\pi^{prompt}_\mathcal{S}\rightarrow\pi_\mathcal{A}$          & 0.682 & 0.585 & 0.663 & 0.046 & 0.180 & 0.699 & 0.495 & 0.660 & 0.049 & 0.280 & 8\% \\
                $\pi^{SFT}_\mathcal{S}\rightarrow\pi_\mathcal{A}$        & 0.721 & 0.600 & 0.667 & 0.025 & 0.180 & 0.715 & 0.545 & 0.710 & 0.044 & 0.365 & 5\% \\
                \hline
                $\pi^{TG}_\mathcal{S}\rightarrow\pi_\mathcal{A}$        & 0.658 & \textbf{0.630} & 0.677 & 0.037 & 0.203 & 0.716 & \textbf{0.595} & 0.677 & 0.019 & \textbf{0.393} & 1\% \\
                $\pi^{DSPy}_\mathcal{S}\rightarrow\pi_\mathcal{A}$      & 0.695 & 0.620 & 0.677 & 0.039 & 0.200 & 0.719 & 0.575 & 0.683 & 0.025 & \textbf{0.393} & 4\% \\
                \hline 
                $\pi^{1}_\mathcal{S}\rightarrow\pi_\mathcal{A}$          & 0.756 & 0.610 & \textbf{0.707} & 0.028 & 0.198 & 0.761 & 0.545 & \textbf{0.717} & 0.048 & 0.338 & 10\% \\
                $\pi^{2}_\mathcal{S}\rightarrow\pi_\mathcal{A}$          & 0.798 & 0.585 & 0.697 & 0.030 & 0.205 & 0.746 & 0.510 & 0.707 & 0.048 & 0.338 & 10\% \\
                $\pi^{3}_\mathcal{S}\rightarrow\pi_\mathcal{A}$          & 0.810 & 0.590 & 0.703 & \textbf{0.048} & 0.205 & \textbf{0.761} & 0.565 & 0.707 & 0.042 & 0.385 & 16\% \\
                $\pi^{4}_\mathcal{S}\rightarrow\pi_\mathcal{A}$          & \textbf{0.825} & 0.620 & 0.700 & 0.037 & 0.212 & 0.747 & 0.520 & 0.713 & 0.046 & 0.385 & 14\% \\
                $\pi^{5}_\mathcal{S}\rightarrow\pi_\mathcal{A}$          & 0.823 & 0.615 & 0.703 & 0.044 & 0.242 & 0.744 & 0.570 & 0.707 & \textbf{0.053} & 0.385 & \textbf{21\%} \\
                \bottomrule
            \end{tabular}
        }
    }
    \caption{Performance comparison on five benchmarks under two actors: v3.5-sonnet-v2 and gpt-oss-120b. Supplement is generated by Qwen3-1.7B model. The best results are highlighted in bold.}
    \label{tbl:main}
\end{table*}

\noindent\textbf{\\\textit{Baselines:}}

    • $\emptyset\rightarrow\pi_\mathcal{A}$: Baseline performance of an actor model $\pi_\mathcal{A}$ without using supplements.
    
    • $\emptyset\rightarrow\pi^{ITS}_\mathcal{A}$: Performance of $\pi_\mathcal{A}$ with inference time scaling, prompted with chain-of-thought \cite{wei2022chainofthought}.
    
    • $\emptyset\rightarrow\pi^{solve}_\mathcal{S}$: Performance of a supplement generator model $\pi_\mathcal{S}$, but SFT trained to solve the task instead of generating supplements.
    
    • $\pi^{prompt}_\mathcal{S}\rightarrow\pi_\mathcal{A}$: Performance of $\pi_\mathcal{A}$, with untrained $\pi_\mathcal{S}$ prompted with \textit{Free Style} supplement type prompt.
    
    • $\pi^{SFT}_\mathcal{S}\rightarrow\pi_\mathcal{A}$: Performance of $\pi_\mathcal{A}$, with $\pi_\mathcal{S}$ only trained with SFT.
    
     • $\pi^{TG}_\mathcal{S}\rightarrow\pi_\mathcal{A}$ \cite{yuksekgonul2025textgrad}: We use TextGrad to automatically optimize the textual variables of the supplement generator. TextGrad treats feedback generated by LLM as textual gradients and applies a gradient-descent-like update loop to iteratively refine the prompt such that the downstream metric improves. Then the optimized prompts are provided to $\pi_\mathcal{A}$.

    • $\pi^{DSPy}_\mathcal{S}\rightarrow\pi_\mathcal{A}$ \cite{khattab2023dspy}: We implement the supplement generation procedure as a DSPy program, which is a declarative module mapping the task input to a supplement, and then use DSPy’s compiler algorithms to automatically learn the prompt parameters that maximize the downstream metric. The compiled module $\pi^{DSPy}_\mathcal{S}$ is then used to generate supplements at inference time, which are passed to $\pi_\mathcal{A}$.

\noindent\textbf{\\\textit{Proposed Methods:}}

    • $\pi^{*}_\mathcal{S}\rightarrow\pi_\mathcal{A}$: Performance of $\pi_\mathcal{A}$, with DPO trained $\pi_\mathcal{S}$ , showing the best performance in $T$ iterations.
    
    • $\pi^{t}_\mathcal{S}\rightarrow\pi_\mathcal{A}$: Performance of $\pi_\mathcal{A}$, with DPO trained $\pi_\mathcal{S}$ , showing the performance in $t^{th}$ iteration.

\subsection{Foundation Models}
In our experiments, we use the Qwen3-1.7B model for $\pi_\mathcal{S}$ with the thinking mode disabled \cite{yang2025qwen3}. For $\pi_\mathcal{A}$, we employ the v3.5-sonnet-v2 and GPT-OSS-120B models, with the reasoning effort of GPT-OSS-120B set to medium \cite{anthropic2024sonnet, openai2025gptoss120bgptoss20bmodel}. The Qwen3-1.7B model is used to highlight the capability of a compact model to generate supplements, while Sonnet and GPT-OSS are utilized to compare the effects of thinking versus non-thinking configurations. The three models are deliberately chosen from different model families to ensure that our findings generalize across diverse architectural and training paradigms.

\section{Results}

This section reports the empirical performance of our proposed method, comparing it with a comprehensive suite of baselines and ablations. We analyze (1) overall performance gains, (2) iteration-wise improvements during DPO training, and (3) task-specific trends and robustness across models.

\begin{figure}[t]
    \centering
    \includegraphics[width=\linewidth]{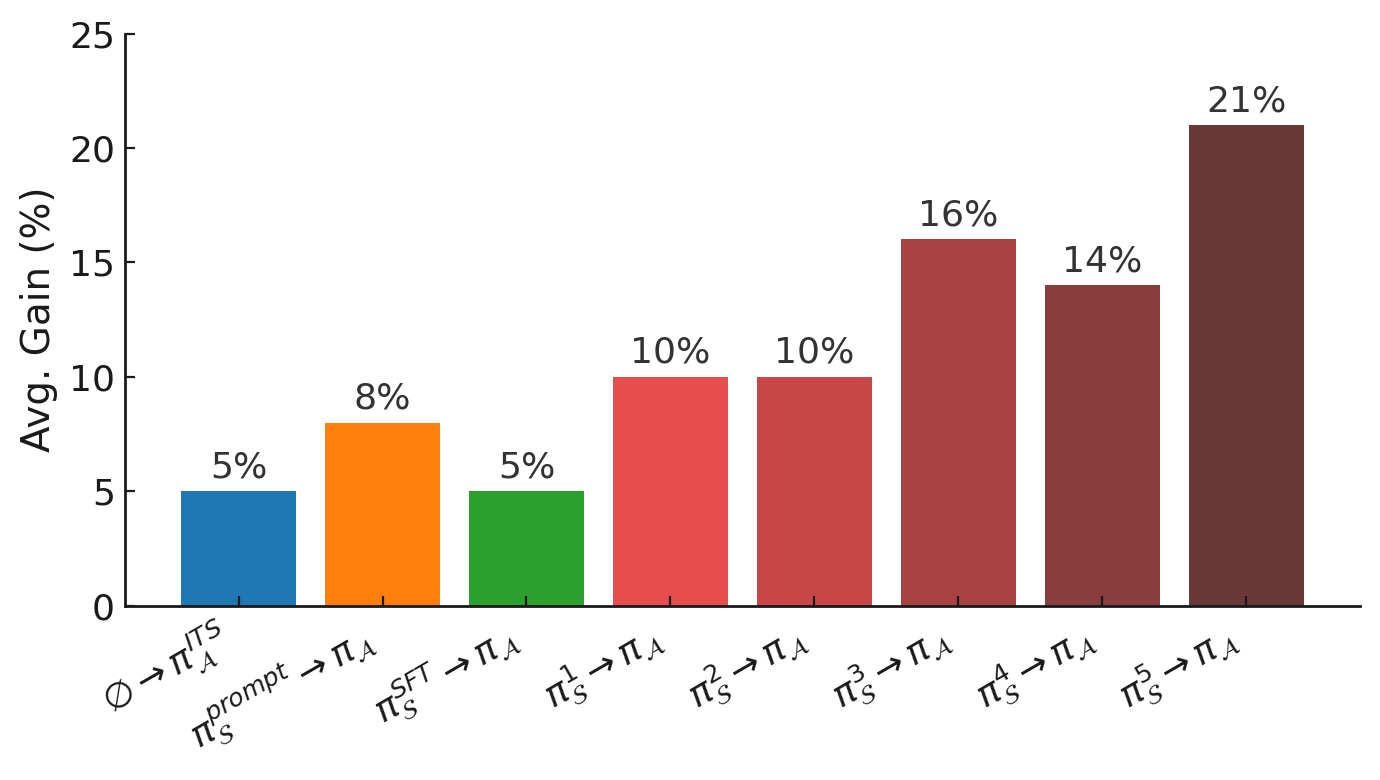}
    \caption{Average performance gain across different methods. Baseline methods are shown in distinct colors, while DPO iterations (1–5) use progressively darker red tones to indicate training progression.}
    \label{fig:barchart}
\end{figure}

\begin{figure*}[t]
    \centering
    \includegraphics[width=\textwidth]{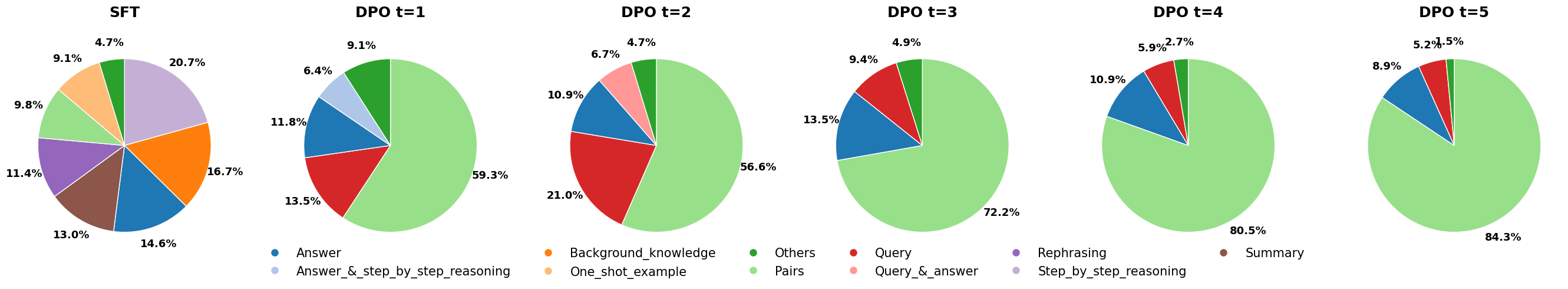}
    \caption{Distribution shift of generated supplement types across training stages. Showing the example of Qwen3-1.7B generates supplement to GPT-OSS-120B model on Spider benchmark.}
    \label{fig:pies}
\end{figure*}

\begin{figure*}[t]
    \centering
    \includegraphics[width=\textwidth]{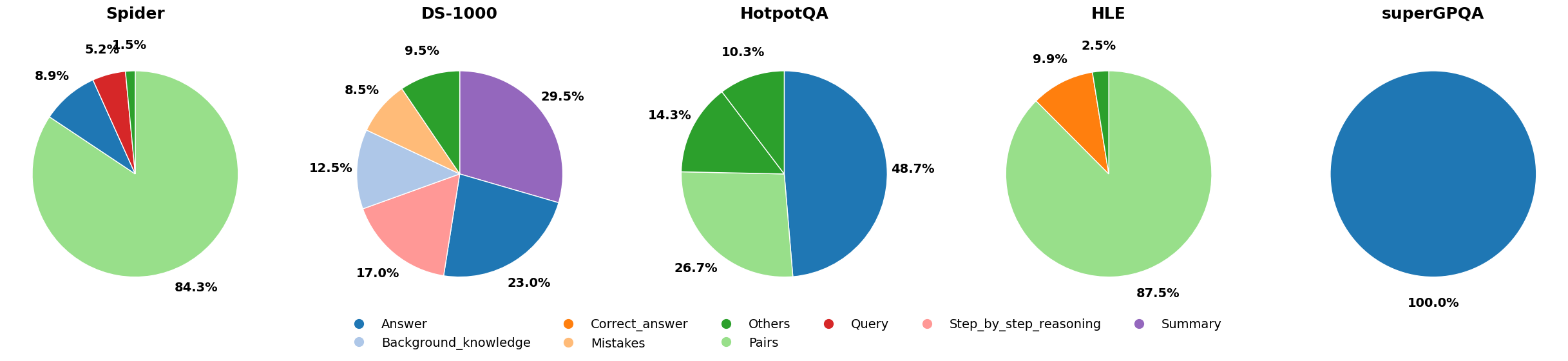}
    \caption{Distribution of generated supplement types across different benchmarks. The figure illustrates how the model adapts its supplement generation strategy to task characteristics.}
    \label{fig:benchmarks}
\end{figure*}

\subsection{Overall Comparison with Baselines}

As shown in Table~\ref{tbl:main} and Figure \ref{fig:barchart}, the proposed approach ($\pi^{5}_{\mathcal{S}}\!\rightarrow\!\pi_{\mathcal{A}}$) achieves the \textbf{highest overall performance} across all benchmarks, with an \textbf{average gain of 21\%} over the baseline actor $\emptyset\!\rightarrow\!\pi_{\mathcal{A}}$. Performance increases steadily across DPO training iterations ($\pi^{t}_{\mathcal{S}}\!\rightarrow\!\pi_{\mathcal{A}}$, $t=1\dots5$). Early iterations already outperform all baselines: +10\% gain at $t=1$, and +21\% by $t=5$. This demonstrates that integrating DPO-trained supplement generators consistently improves the downstream reasoning and task performance of the actor models.

Compared to inference-time scaling ($\emptyset\!\rightarrow\!\pi^{ITS}_{\mathcal{A}}$), our method yields \textbf{substantial additional improvement} (from +5\% to +21\%), indicating that dynamic supplement generation offers stronger benefits than chain-of-thought prompting alone. Likewise, while SFT-only training ($\pi^{SFT}_{\mathcal{S}}\!\rightarrow\!\pi_{\mathcal{A}}$) provides moderate improvements (+5\%), further fine-tuning with DPO yields an \textbf{additional 16 percentage points} in relative gain, confirming the importance of preference-based optimization.

Our method demonstrates superior performance compared to two popular APO methods, TextGrad and DSPy. Across most experimental settings, SGT consistently achieves higher accuracy or performs on par with the baselines, while the two APO methods exhibit more inconsistent performance. On average, the fifth iteration of DPO with SGT yields a 21\% performance improvement, whereas TextGrad and DSPy show only marginal gains.

\subsection{Task-Specific Observations}

Across benchmarks, we observe that the effectiveness of supplement generation depends on the nature of reasoning demanded by each task. The largest gains appear in structured reasoning tasks such as Spider and DS-1000, where the actor must decompose inputs into compositional logic or code structures. In these settings, supplements likely help by externalizing intermediate reasoning steps and clarifying latent constraints before execution, allowing the actor to plan more systematically. For instance, Spider accuracy increases from 0.641 to 0.825 under v3.5-sonnet-v2, indicating that supplements enable more consistent schema mapping and query formulation.

In contrast, improvements on open-ended reasoning tasks such as HotpotQA are comparatively modest. Because these tasks rely heavily on factual recall and long-context reading rather than compositional inference, the supplement generator’s structured guidance offers limited additional benefit once the actor already retrieves the relevant evidence. This suggests that supplements are most valuable when the bottleneck lies in reasoning organization rather than knowledge access.

On difficult evaluation tasks such as HLE and superGPQA, we see smaller absolute but meaningful relative improvements. These tasks are characterized by ambiguous or high-dimensional reasoning objectives, where small changes in model calibration can have large effects on final success rates. The observed gains imply that DPO training helps the supplement generator produce more targeted, context-aware guidance that mitigates the actor’s uncertainty during inference.

\subsection{Ablation Insights}

Ablation studies further clarify the contribution of each component. When the supplement generator is untrained and only prompted in a “free style” manner ($\pi^{prompt}_{\mathcal{S}}\!\rightarrow\!\pi_{\mathcal{A}}$), performance improves only marginally (+8\%), showing limited benefit from uninformed supplement generation. In contrast, when the model is fine-tuned to directly solve tasks instead of generating supplements ($\emptyset\!\rightarrow\!\pi^{solve}_{\mathcal{S}}$), performance drops dramatically (–23\%), indicating the difficulty of training a small model for task-solving ability, instead of supplement generation. Together, these findings underscore that \textbf{training specifically for supplement generation and optimizing via DPO} is essential for achieving the observed gains.

\section{Qualitative Analysis}
\subsection{Supplement Type Distribution Shift}
\label{sec:diff_shift}
In this experiment, we examine the distribution of supplement types generated by $\pi_\mathcal{S}$ across different training stages, as illustrated in Figure \ref{fig:pies} using the example on Spider benchmark using GPT-OSS-120B. During the SFT stage, the eight pre-defined supplement types are relatively evenly distributed, reflecting the goal of SFT training: to help the model internalize and reproduce all supplement types, and also demonstrating the effectiveness of stratified sampling. In contrast, the distribution after the first iteration of DPO becomes noticeably skewed: the \textit{Pairs} supplement type dominates, and a new type emerges. This shift highlights the dual objectives of DPO training: \textit{search-and-focus}. As DPO progresses through subsequent iterations, the distribution becomes increasingly concentrated around \textit{Pairs}, while still preserving a smaller proportion of other supplement types to allow for instance-specific optimization.

\subsection{Task-Specific Distribution of Generated Supplement Types}
\label{sec:task_shift}
The distribution of generated supplement types across different benchmarks is shown in Figure~\ref{fig:benchmarks}. The composition varies notably by task, revealing dataset-specific preferences in model behavior. For \textit{Spider} and \textit{HLE}, the \textit{Pairs} supplement type overwhelmingly dominates, suggesting that tasks requiring structured or multi-step reasoning tend to favor pairwise comparison supplements. In contrast, \textit{DS-1000} shows a more diverse distribution, with substantial proportions of \textit{Summary} (29.5\%), \textit{Step-by-step reasoning} (17.0\%), and \textit{Answer} (12.5\%), indicating that code-related reasoning tasks elicit a broader range of supplement strategies. \textit{HotpotQA} exhibits a balanced mix between \textit{Answer} and \textit{Pairs}, consistent with its multi-hop reasoning nature, where both direct answers and comparative reasoning are valuable. Finally, \textit{SuperGPQA} exclusively generates \textit{Answer} supplements, suggesting a task formulation that emphasizes precision over reasoning diversity. Collectively, these patterns demonstrate that supplement-type distributions adapt dynamically to task demands, reflecting the model’s capacity to adjust reasoning strategies to fit the problem structure.

\section{Conclusion}
In this work, we introduced Supplement Generation Training, a lightweight and generalizable framework that trains a small language model to generate per-input supplements that enhance the performance of large models. SGT achieves consistent and significant performance gains across diverse tasks and models, all without requiring gradient access to the actor model. The results highlight that supplements can serve as effective, interpretable intermediaries for improving reasoning and robustness in large language models. 

\section*{Limitations}

\paragraph{Dataset Scale.} The datasets used in our experiments ranged from hundreds to thousands of examples. While evaluating the scalability of our method as data sizes increase might be interesting, we intentionally restrict the data size to simulate realistic low-resource settings. 

\paragraph{Supplement Types.} The current work focuses on a predefined set of 8 supplement types. While our method allows the pipeline to be initiated with other set of supplement types and definitions, we ground our selection from previous works, carefully curate essential supplements to represent wide coverage of additional information that can be delivered to the actor model.

\section*{Acknowledgements}
This work was supported in part by AWS Agentic AI Labs and compute resources provided through Amazon. We thank members of the AWS Agentic AI Labs for helpful discussions and feedback. The first author conducted this work during an internship at Amazon. AI writing assistants were used for grammar checking in the preparation of this manuscript.

\bibliography{custom}

\appendix
\section{Supplement Types}
\label{supp_types}
In this section, we show eight pre-defined supplement types, with \textit{Free Style} supplement type, for their definition and prompt used in SFT data construction.

\begin{itemize}
    \item \textbf{Free Style}:
    \begin{itemize}
        \item \textit{Prompt}: “Based on the task above, please provide supplementary text that can assist in completing the task.”
        \item \textit{Goal}: ensure maximum supplement type variability without an instruction.

        \item \textit{Example}:\{"supplementary\_text": \}

    \end{itemize}

    \item \textbf{Answer}:
    \begin{itemize}
        \item \textit{Prompt}: "" (solve the task directly)
        \item \textit{Goal}: provide direct solution.

        \item \textit{Example}: \{"answer": \}

    \end{itemize}

    \item \textbf{Background}:
    \begin{itemize}
        \item \textit{Prompt}: “Based on the task above, please provide background knowledge that does not exist in the task.”

        \item \textit{Goal}: provide external information, or further explanation to the input.

        \item \textit{Example}: \{"background\_knowledge": \}

    \end{itemize}

    \item \textbf{CoT}:
    \begin{itemize}
        \item \textit{Prompt}: “Based on the task above, please provide a step by step reasoning that can assist in completing the task.”
        \item \textit{Goal}: provide step by step reasoning (planning) to the task.
        \item \textit{Example}: \{"step\_by\_step\_reasoning":\} 
    \end{itemize}

    \item \textbf{Rephrase}:
    \begin{itemize}
        \item \textit{Prompt}: “Based on the task above, please rephrase the task to make it clearer and more understandable.”

        \item \textit{Goal}: reduce task ambiguity.
        \item \textit{Example}: \{"rephrasing": \}
    \end{itemize}

    \item \textbf{Summary}:
    \begin{itemize}
        \item \textit{Prompt}: “Based on the task above, please first provide a summary of context (excluding the specific question).”
        \item \textit{Goal}: reducing actor model’s cognitive load by summarizing context.
        \item \textit{Example}: \{"summary": \}
    \end{itemize}

    \item \textbf{Mistakes}:
    \begin{itemize}
        \item \textit{Prompt}: “Based on the task above, please provide common mistakes in completing the task.”
        \item \textit{Goal}: avoid making common and simple mistakes.
        \item \textit{Example}: \{"mistakes": \}
    \end{itemize}

    \item \textbf{One-Shot}:
    \begin{itemize}
        \item \textit{Prompt}: “Based on the task above, please provide one different question+answer example”
        \item \textit{Goal}: synthetic generation of one-shot example.
        \item \textit{Example}: \{"one\_shot\_example": \}
    \end{itemize}

    \item \textbf{Pairs}:
    \begin{itemize}
        \item \textit{Prompt}: “Following the answer format above, please provide a correct answer and an incorrect answer to this task that illustrates common mistakes.”
        \item \textit{Goal}: provide correct and incorrect answer pairs to infer mistakes to avoid.
        \item \textit{Example}: \{"correct\_answer":, "incorrect\_answer":\} 

    \end{itemize}
    
\end{itemize}

\section{Dataset Selection}

We show the basic statistics of the benchmarks in Table \ref{tab:benchmark_sizes}. In this experiment, we used Spider, DS-1000, and HotpotQA from StreamBench evaluation pipeline and its samples \cite{wu2024streambench}. For HLE, we used text-only questions; for superGPQA, we randomly sampled 2000 examples from hard questions. We further separated benchmarks into training, validation, and test set using 6:2:2 ratio. We used training set only for SFT training, and validation set is used only for DPO training. All performances and results are reported with test splits. During training, we used gold answers and examples to construct \textit{Answer}, \textit{Pairs}, and \textit{One-Shot} supplement types.

\begin{table}[h]
\centering

\begin{tabular}{l c}
\toprule
\textbf{Benchmark} & \textbf{Data Size} \\
\midrule
Spider & 2,147 \\
DS-1000 & 1,000 \\
HotpotQA & 1,500 \\
HLE & 2,158 \\
superGPQA & 2,000 \\
\bottomrule
\end{tabular}
\caption{Data sizes of benchmarks used in evaluation.}
\label{tab:benchmark_sizes}
\end{table}

\section{Advantages over Existing Prompt Optimization Methods}
Compared to existing Automatic Prompt Optimization and Black-Box Prompt Optimization methods, SGT introduces several conceptual and practical advantages:
\begin{enumerate}
    \item Per-Instance Adaptation: Unlike global template-based APO methods, SGT dynamically tailors supplements to each input, allowing fine-grained reasoning control.
    \item No Gradient Access Required: SGT leverages proxy rewards, enabling training with API-only actor models.
    \item Lightweight and Modular: The supplement generator can be updated or replaced independently of the actor model, supporting sustainable iteration without full retraining.
    \item Interpretability: Supplements are human-readable, offering transparency into the reasoning augmentation process.
\end{enumerate}

These features collectively make SGT a promising framework for real-world deployments where model retraining is infeasible or governance constraints prohibit weight updates.

\section{Ablation on DPO Iterations}
\label{app:dpo_iter}
In the main experiments, we altered the number of DPO iterations from 1 to 5 to show how model’s performance increase, also how the distribution of supplement types change round by round. In this section, we additionally increased the number of DPO iteration to 10. The result is presented in Table \ref{tbl:dpo10}.

The result presents that the best average score is achieved at 5th DPO, where prolonged training makes performance inconsistent. While some domains such as HotpotQA benefit from extended training iterations, other benchmarks show fluctuating performance after 6th iteration, indicates overfitting.

\section{Ablation on Model Sizes}
\label{app:model_size}
We employ Qwen3-1.7B model for our main experiments, which presented a strong performance within a highly cost-efficient setting. In this section, we provide additional experiment results where a supplement generator model Qwen3-4B is paired with actor model gpt-oss-120b. The result is presented in Table \ref{tbl:model_ablation}. With Qwen3-4B model, SGT still shows a strong performance increase, higher than all baselines including SFT.

\begin{table}[t!]
\centering
\resizebox{\linewidth}{!}{
\begin{tabular}{lcccccc}
\toprule
Method & Spider & DS-1000 & HotpotQA & HLE & SuperGPQA & Avg. Gain \\
\midrule
$\emptyset \rightarrow \pi_A$ 
& 0.641 & 0.565 & 0.690 & 0.035 & 0.203 & - \\

$\emptyset \rightarrow \pi^{\text{ITS}}_A$ 
& 0.647 & 0.580 & 0.647 & 0.019 & 0.295 & -1\% \\

$\pi^{\text{SFT}}_S \rightarrow \pi_A$ 
& 0.744 & 0.585 & 0.690 & 0.028 & 0.328 & 12\% \\

$\pi^1_S \rightarrow \pi_A$ 
& 0.741 & 0.555 & 0.723 & 0.044 & 0.347 & 23\% \\

$\pi^2_S \rightarrow \pi_A$ 
& 0.736 & 0.570 & 0.700 & 0.048 & 0.383 & 29\% \\

$\pi^3_S \rightarrow \pi_A$ 
& 0.734 & 0.570 & 0.700 & 0.039 & 0.378 & 23\% \\

$\pi^4_S \rightarrow \pi_A$ 
& 0.747 & 0.580 & 0.730 & 0.039 & 0.378 & 25\% \\

$\pi^5_S \rightarrow \pi_A$ 
& 0.761 & 0.605 & 0.693 & 0.046 & 0.365 & 27\% \\
\bottomrule
\end{tabular}
}

\caption{Performance comparison across datasets with Qwen3-4B model as the supplement generator model. GPT-OSS-120B is used as the actor model.}
\label{tbl:model_ablation}
\end{table}

\begin{table}[t!]
\centering
\resizebox{\linewidth}{!}{
\begin{tabular}{llll}
\toprule
       & $\emptyset \rightarrow \pi_A$     & $\pi^*_S \rightarrow \pi_A$ (trained with BIRD)      & $\pi^*_S \rightarrow \pi_A$ (trained with Spider)      \\
\hline
BIRD   & 0.361 & 0.451 & 0.376 \\
Spider & 0.641 & 0.730 & 0.828 \\
\bottomrule
\end{tabular}
}
\caption{Result of the cross-dataset experiment with Qwen3-1.7B model as supplement generator model, and v3.5-sonnet-v2 model as the actor model.}
\label{tbl:cross_dataset}
\end{table}

\section{Ablation on Cross Dataset Performance}
\label{app:cross_dataset}
Because our SGT method is designed to be lightweight, efficient, and easy to train or apply across new benchmarks, cross-benchmark generalizability is not the main emphasis of this work. Regardless, we present an extra experiment where the SGT method is trained on Spider and evaluated on BIRD \cite{li2023bird}, and vice versa. The result is presented in Table \ref{tbl:cross_dataset}. When the SGT trained model is used on other benchmarks, it presented stronger performance than the baseline (0.361 -> 0.376, 0.641 -> 0.730), showing the transferability of knowledge from one benchmark to another. Interestingly, the performance increase is stronger when the model is trained on a harder dataset, and applied to an easier one.

\section{Supplement Type Distribution}
In Figure \ref{fig:full_piechart_1.7b}, we present the shift and convergence of supplement type distributions across DPO iterations and benchmarks. Across all five experiments, the distributions after SFT are highly diverse and quasi-uniform, reflecting balanced exposure to all predefined types. As DPO iterations progress, new supplement types emerge and the distributions gradually concentrate toward a small number of dominant types, adapting to the specific characteristics of each benchmark.

\begin{figure*}[t]
    \centering
    \includegraphics[width=\linewidth]{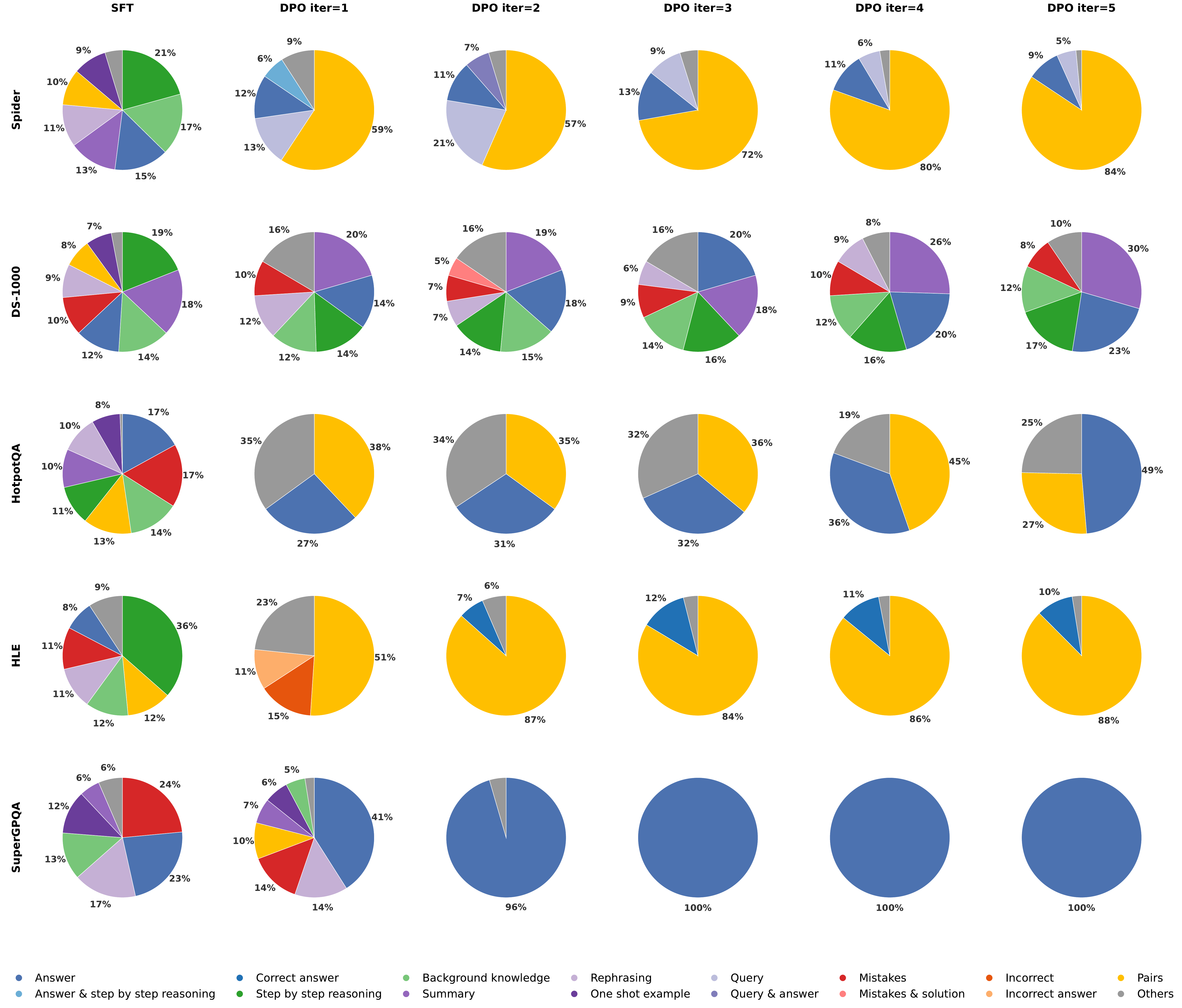}
    \caption{Distribution of generated supplement types across training stages and benchmarks. After SFT, all benchmarks exhibit a quasi-uniform distribution across predefined types. As DPO iterations progress, new types emerge and the distributions converge toward a small number of dominant types tailored to each benchmark: \textit{Pairs} for Spider, \textit{Answer} for SuperGPQA, and mixed strategies for DS-1000 and HotpotQA. The supplement generator is Qwen3-1.7B and the actor model is GPT-OSS-120B.}
    \label{fig:full_piechart_1.7b}
\end{figure*}

\section{Latency Overhead and Cost Analysis}
We used an EC2 p4d instance for our experiments. The average length of the input tokens is 220, and output token length is 75. Running a Qwen3-1.7B based model on our machine, generating a supplement only takes 15-30 milliseconds per request. Based on most LLM APIs respond in 150-500 ms for small outputs, SGT will not be a bottleneck for most use cases. Training a Qwen3-1.7B model on our machine takes about 1.5 hours in average, where SFT takes about 30 minutes and DPO takes about an hour.

Together, short generation latency and small training cost emphasize our main contribution: cost reduction.

\begin{table*}[t!]
    \centering
    \resizebox{\textwidth}{!}{
        {\renewcommand{\arraystretch}{1.3}
            \begin{tabular}{lccccccccccc}
                \toprule
                \multirow{2}{*}{\textbf{Method}} & \multicolumn{5}{c}{\textbf{v3.5-sonnet-v2}} & \multicolumn{5}{c}{\textbf{gpt-oss-120b}} & \multirow{2}{*}{\textbf{Avg. Gain}} \\
                \cmidrule(lr){2-6} \cmidrule(lr){7-11}
                & \textbf{Spider} & \textbf{DS-1000} & \textbf{HotpotQA} & \textbf{HLE} & \textbf{superGPQA} & \textbf{Spider} & \textbf{DS-1000} & \textbf{HotpotQA} & \textbf{HLE} & \textbf{superGPQA} & \\
                \midrule
                $\pi^{1}_\mathcal{S}\rightarrow\pi_\mathcal{A}$          & 0.756 & 0.610 & 0.707 & 0.028 & 0.198 & 0.761 & 0.545 & 0.717 & 0.048 & 0.338 & 10\% \\
                $\pi^{2}_\mathcal{S}\rightarrow\pi_\mathcal{A}$          & 0.798 & 0.585 & 0.697 & 0.030 & 0.205 & 0.746 & 0.510 & 0.707 & 0.048 & 0.338 & 10\% \\
                $\pi^{3}_\mathcal{S}\rightarrow\pi_\mathcal{A}$          & 0.810 & 0.590 & 0.703 & \textbf{0.048} & 0.205 & 0.761 & 0.565 & 0.707 & 0.042 & 0.385 & 16\% \\
                $\pi^{4}_\mathcal{S}\rightarrow\pi_\mathcal{A}$          & 0.825 & \textbf{0.620} & 0.700 & 0.037 & 0.212 & 0.747 & 0.520 & 0.713 & 0.046 & 0.385 & 14\% \\
                $\pi^{5}_\mathcal{S}\rightarrow\pi_\mathcal{A}$          & 0.823 & 0.615 & 0.703 & 0.044 & \textbf{0.242} & 0.744 & \textbf{0.570} & 0.707 & \textbf{0.053} & 0.385 & \textbf{21\%} \\
                $\pi^{6}_\mathcal{S}\rightarrow\pi_\mathcal{A}$         & 0.822 & 0.585 & 0.707 & 0.044 & 0.21  & 0.774 & 0.53  & 0.703 & 0.042 & \textbf{0.388} & 14\% \\
                $\pi^{7}_\mathcal{S}\rightarrow\pi_\mathcal{A}$         & \textbf{0.828} & 0.575 & 0.71  & 0.046 & 0.237 & \textbf{0.769} & 0.545 & 0.713 & 0.044 & 0.357 & 17\% \\
                $\pi^{8}_\mathcal{S}\rightarrow\pi_\mathcal{A}$         & 0.815 & 0.575 & \textbf{0.713} & 0.039 & 0.233 & 0.764 & 0.55  & 0.727 & 0.035 & 0.347 & 11\% \\
                $\pi^{9}_\mathcal{S}\rightarrow\pi_\mathcal{A}$         & 0.83  & 0.585 & \textbf{0.713} & 0.044 & 0.212 & 0.756 & 0.535 & 0.72  & 0.046 & 0.38  & 16\% \\
                $\pi^{10}_\mathcal{S}\rightarrow\pi_\mathcal{A}$        & 0.823 & 0.59  & 0.703 & 0.035 & 0.24  & 0.751 & 0.545 & \textbf{0.733} & 0.044 & 0.372 & 14\% \\
                \bottomrule
            \end{tabular}
        }
    }
    \caption{Performance comparison on DPO iterations under two actors: v3.5-sonnet-v2 and gpt-oss-120b. Supplement is generated by Qwen3-1.7B model. The best results are highlighted in bold.}
    \label{tbl:dpo10}
\end{table*}

\end{document}